\documentclass{article}

\usepackage{arxiv}
\usepackage{tabu}
\usepackage[utf8]{inputenc} 
\usepackage[T1]{fontenc}    
\usepackage{hyperref}       
\usepackage{url}            
\usepackage{booktabs}       
\usepackage{amsfonts}       
\usepackage{nicefrac}       
\usepackage{microtype}      
\usepackage{lipsum}		
\usepackage{enumitem}
\usepackage{authblk}

\usepackage{graphicx}
\usepackage{amsmath}
\hypersetup{final}

\title{DurIAN: Duration Informed Attention Network For Multimodal Synthesis}

\author{Chengzhu~Yu}
\author{Heng~Lu}
\author{Na~Hu}
\author{Meng~Yu}
\author{Chao~Weng}
\author{Kun~Xu}
\author{Peng~Liu}
\author{Deyi~Tuo}
\author{Shiyin~Kang}
\author{Guangzhi~Lei}
\author{Dan~Su}
\author{Dong~Yu}
\affil{Tencent AI Lab}

\begin{document}
\maketitle

\begin{abstract}
In this paper, we present a generic and robust multimodal synthesis system that produces highly natural speech and facial expression simultaneously. The key component of this system is the Duration Informed Attention Network (DurIAN), an autoregressive model in which the alignments between the input text and the output acoustic features are inferred from a duration model. This is different from the end-to-end attention mechanism used, and accounts for various unavoidable artifacts, in existing end-to-end speech synthesis systems such as Tacotron. Furthermore, DurIAN can be used to generate high quality facial expression which can be synchronized with generated speech with/without parallel speech and face data. To improve the efficiency of speech generation, we also propose a multi-band parallel generation strategy on top of the WaveRNN model. The proposed Multi-band WaveRNN effectively reduces the total computational complexity from 9.8 to 3.6 GFLOPS, and is able to generate audio that is 6 times faster than real time on a single CPU core. We show that DurIAN could generate highly natural speech that is on par with current state of the art end-to-end systems, while at the same time avoid word skipping/repeating errors in those systems. Finally, a simple yet effective approach for fine-grained control of expressiveness of speech and facial expression is introduced. 
\end{abstract}


\section{Introduction}
Text-to-speech synthesis (TTS) is the task of converting source texts into speech signals that sound like natural human speech. The quality of speech synthesis systems is evaluated based on multiple factors including naturalness, robustness, and accuracy of generated speech. For many real-world speech synthesis tasks, generation time, latency, and computational costs are also important factors to consider. Recently, there is increasing demand for generating multimodal signals which requires the generated speech and facial expression being both natural and synchronized. Combined with facial modeling techniques, the multimodal synthesis techniques make it possible to achieve the goal of natural face-to-face communication in many human-computer interaction scenarios such as virtual worlds, computer games, online education and so on. 


Traditional speech synthesis approaches, including concatenative methods \cite{hunt1996unit,black1997automatically} and statistic parametric systems \cite{tokuda2000speech,zen2009statistical,ze2013statistical}, are all based on acoustic feature analysis and synthesis. These approaches are still predominantly used in industrial applications due to their advantages in robustness and efficiency. However, these approaches suffer from the inferior naturalness of generated speech. End-to-end approaches \cite{wang2017tacotron,shen2018natural,li2018close,ping2018clarinet,sotelo2017char2wav,ping2017deep} have gained much attention recently due to the clearly better naturalness of their synthesized results and the simplified training pipelines. 
Unfortunately, the existing end-to-end systems are lack of robustness when generating speech as they  produce unpredictable artifacts where random words in the source text are repeated or skipped in generated speech \cite{shen2018natural,ping2017deep} esp. when synthesizing out-of-domain texts. 
For multimodal synthesis tasks, synchronization between speech and facial expression is another challenge for end-to-end based systems. While speech and face features can be generated as a pair with end-to-end models, this approach requires large amount of paired speech and facial expression data for training. Such paired speech and facial expression data is expensive to collect and cannot be obtained in scenarios when the desired voice and virtual image come from different sources.

In this paper, we propose duration informed attention network (DurIAN), a generic multimodal synthesis framework that generates highly natural and robust speech and facial expression\footnote{Sound and video demo can be found at https://tencent-ailab.github.io/durian/}. DurIAN is a combination of traditional parametric systems and recent end-to-end systems, and can achieve both naturalness and robustness in speech generation. The recent end-to-end systems advance over the traditional parametric systems in multiple perspectives, including the use of an encoder to replace the manually-designed linguistic features, an autoregressive model to address the prediction oversmoothing problem, a neural vocoder to replace the traditional source-filter vocoder, and an attention mechanism for end-to-end training and optimization. Our observation and analysis indicates that the speech generation instability in existing end-to-end systems is caused by the end-to-end attention mechanism. Therefore, the core idea behind DurIAN is to replace the end-to-end attention mechanism with an alignment model similar to the one in parametric systems, while preserving other advancements in existing end-to-end systems \footnote{At the time of preparing this paper, we became aware of a preprint paper \cite{ren2019fastspeech} where a similar idea was proposed to address the robustness issues related to the end-to-end systems. Our work is independently developed as indicated by the patent \cite{DurianPatent} filed before their preprint paper and the fact that many design choices are completely different.}. Moreover, the alignment model includes a duration prediction model which can be used for driving facial expression generation without relying on parallel speech and face data.

\begin{figure}[t!]
 \centering
  \includegraphics[clip, trim=0.2cm 9cm 0.2cm 9cm, width=1.00\textwidth]{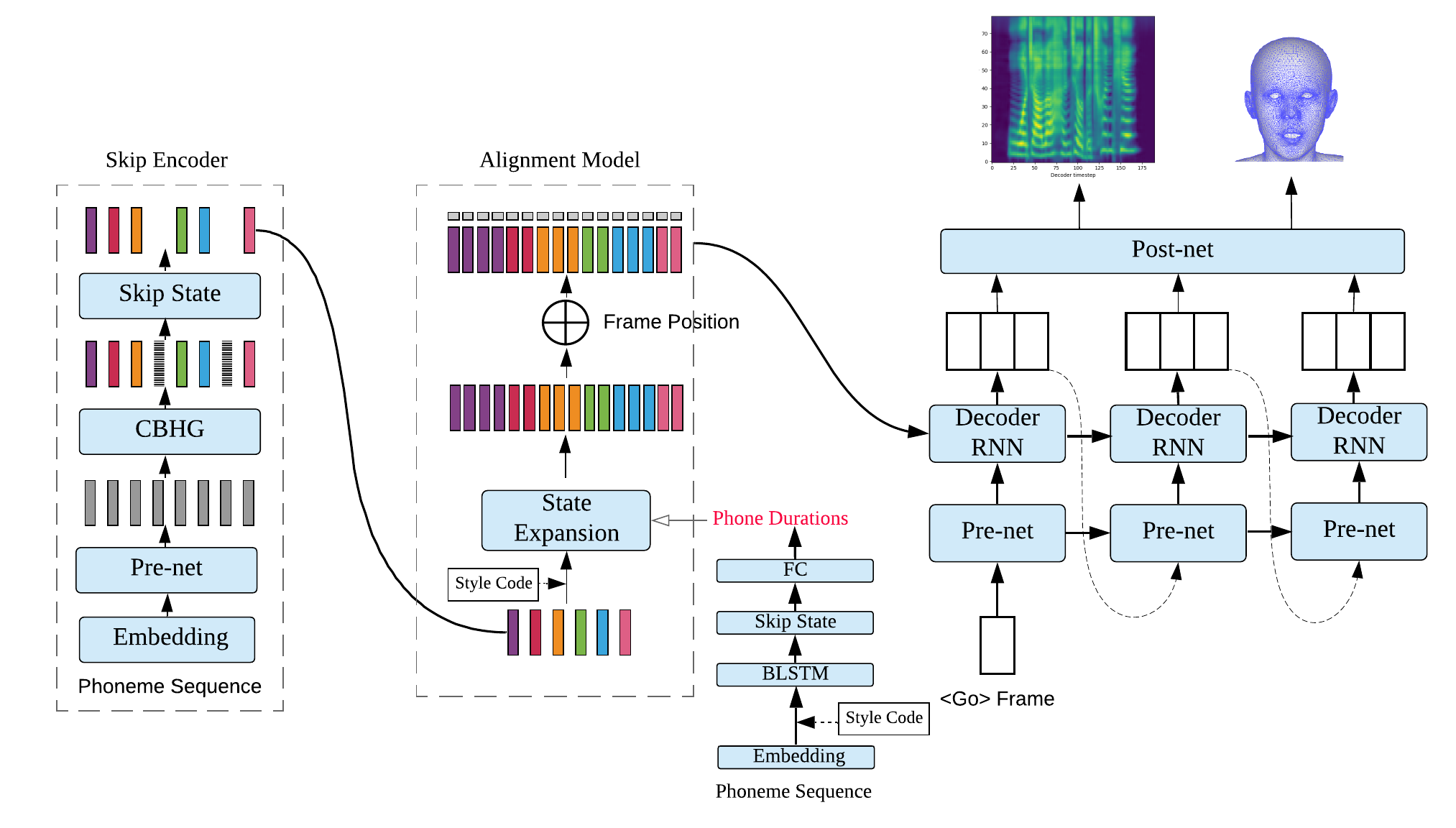}
 \caption{Model architecture of DurIAN. The model takes a sequence of symbols, including phonemes and prosodic boundaries between them, and outputs the corresponding mel-spectrogram and facial expression control parameters.}
 \label{fig:DurIAN}
 \end{figure}
 
The main contributions of this paper are as follows:
\begin{enumerate}
  \item We propose to replace the end-to-end attention mechanism in the Tacotron~2 \cite{shen2018natural} system with the alignment model in traditional parametric systems. We empirically show that the proposed method could generate highly natural speech that is on par with that generated using Tacotron~2, while at the same time DurIAN generated speech is much more robust and stable.
  \item We use a skip encoder structure to encode both the phoneme sequence representation and the hierarchical prosodic structures in Chinese prosody for improved generalization of the DurIAN system on out-of-domain texts in Chinese speech synthesis tasks. 
  \item We propose a simple yet effective fine-grained style-control approach under the supervised setup without fine-grained labels during training as an extension to the conventional multi-style training. 
  \item We describe a multi-band synchronized parallel WaveRNN algorithm to reduce the computational cost in the original WaveRNN model \cite{kalchbrenner2018efficient} and speedup the inference process on single CPU. 
\end{enumerate}

\section{DurIAN}
\label{sec:durian}
In this section, we describe the main components in the DurIAN multimodal synthesis system. As DurIAN is a text driven system, it takes a sequence of symbols converted from text and outputs mel-spectrogram or facial modeling parameters. 

The architecture of DurIAN is illustrated in Figure.~\ref{fig:DurIAN}. It includes (1) a skip encoder that encodes both phoneme sequences and prosodic structures, (2) an alignment model that aligns the input phoneme sequence and the target acoustic frames at frame level, (3) an autoregressive decoder network that generates target acoustic or facial modeling features frame by frame, (4) a \textit{post-net} \cite{shen2018natural} that predicts the residuals not captured by the decoder network. 

The skip encoder takes a sequence of symbols $\mathrm{x_{1:N}}$ as input and outputs a sequence of hidden states $\mathrm{h_{1:N'}}$ as
\begin{equation}
\mathrm{h_{1:N'}} = \mathrm{skip\_encoder}(\mathrm{x_{1:N}}),
\end{equation}
where $\mathrm{N}$ is the length of the input sequence which contains both the phoneme sequence and the prosodic boundaries between them, and $\mathrm{N'}$ is the length of the input phoneme sequence without prosodic boundaries. The length $\mathrm{N}'$ of hidden states output from the skip encoder is different from the length $\mathrm{N}$ of the input sequence because the hidden states associated with the prosodic boundaries are excluded from the final output of the skip encoder (see Sec.~\ref{sec:skip_encoder}). The sequence of hidden states generated from the skip encoder will be expanded according to the duration of each phoneme $\mathrm{d_{1:N'}}$ in the alignment model to generate frame aligned hidden states $\mathrm{e_{1:T}}$ as 

\begin{equation}
\mathrm{e_{1:T}} = \mathrm{state\_expand}(\mathrm{h_{1:N'}}, \mathrm{d_{1:N'}}), 
\end{equation}

where $\mathrm{T}$ equals to the total number of acoustic frames. The state expansion here is basically the replication of hidden states sequentially according to the duration of the given phoneme sequence. During training, the duration of each phoneme is obtained through forced alignment given the input phoneme sequence and the target acoustic features $\mathrm{y_{1:T}}$. 
At the synthesis stage, we exploit the duration of phonemes predicted from the duration model. The expanded hidden states from the alignment model can be exactly paired with the target acoustic frames for training the decoder network to predict each acoustic frame autogressively as 

\begin{equation}
\mathrm{y'_{1:T}} = \mathrm{decoder}(\mathrm{e_{1:T}}),
\end{equation}

where $\mathrm{y'_{1:T}}$ is the predicted acoustic features from decoder network.
The output from the decoder network is passed through a \textit{post-net} to predict the residuals $\mathrm{r_{1:T}}$ as 

\begin{equation}
\mathrm{r_{1:T}} = \mathrm{post}\textrm{-}\mathrm{net}(\mathrm{y'_{1:T}}).
\end{equation}

The entire network is trained to minimize the $\mathrm{\ell1}$ loss 

\begin{equation}
\mathrm{L} = \sum_{n=1}^{T} |\mathrm{y-y'}| +  \sum_{n=1}^{T} |\mathrm{y-(y'+r)}|
\end{equation}

between the predicted and reference mel-spectrograms before and after the \textit{post-net}.

The duration model is separately trained to minimize the $\mathrm{\ell2}$ loss between the predicted and reference duration obtained from forced alignment. In the sections below, we will give a detailed description of the skip encoder, alignment model, and decoder network. As the \textit{post-net} used in DurIAN is exactly the same as the one in Tacotron~2 \cite{shen2018natural}, we will not give detailed description.

\subsection{Skip Encoder}
\label{sec:skip_encoder}
The main objective of the skip encoder is to encode the representation of phoneme sequences as well as hierarchical prosody structure in the hidden states. The prosodic structure is an important component for improved generalization of speech synthesis system on out-of-domain text in Chinese speech synthesis tasks.  

\begin{figure}[h!]
 \centering
  \includegraphics[clip, trim=0.2cm 12cm 0.2cm 12cm, width=0.75\textwidth]{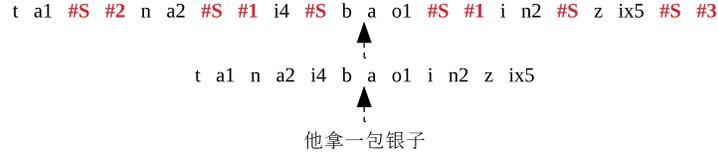}
 \caption{An illustration of how prosodic boundaries are inserted between input phonemes. The symbol \textit{\#S} represents the boundary of syllables, \textit{\#1} represents the boundary of prosodic words, \textit{\#2} represents the boundary of prosodic phrase, and \textit{\#3} represents the boundary of intonational phrase.}
 \label{fig:prosody}
 \end{figure}

To generate the input to the skip encoder, the source text is first converted to a sequence of phonemes. To encode different levels of prosody structures, we insert special symbols representing different levels of prosody boundaries between input phonemes.  Figure.~\ref{fig:prosody} illustrates an example how these special symbols representing prosodic boundaries are inserted.   

The main network components in the skip encoder is inherited from the encoder in the Tacotron~1 \cite{wang2017tacotron} system. Each phoneme and inserted prosodic symbol in the input phoneme sequence is first converted to a continuous vector in the embedding space. The embedded representation of the phoneme sequence is then passed through the \textit{pre-net} \cite{wang2017tacotron} that contains two fully connected layers followed by the CBHG \cite{wang2017tacotron} module. Dropout with probability of 0.5 is applied on the \textit{pre-net} during training. The output from the CBHG module is a sequence of hidden states containing the sequential representation of the input text. Since the alignments between the encoder and decoder states rely on the phone duration model, and since the prosodic boundaries are physically corresponding to time points instead of duration, the hidden states associated with prosodic boundaries are excluded from the output of the CBHG model. An alternative approach for encoding prosodic boundaries is to convert the phoneme sequence into linguistic features where prosodic structures are encoded. However, our early experiments show that using skip encoder could generate speech that is more natural than using linguistic features.

\subsection{Alignment Model}
\label{sec:ali}
One important task in speech synthesis is uncovering the hidden alignment between the phoneme sequence and the target feature/spectrum sequence. End-to-end systems rely on attention based mechanism to discover such alignment. However, existing end-to-end attention mechanism frequently generates unpredictable artifacts where some words are skipped or repeated in the generated speech. Since production speech synthesis systems have very low tolerance on such instability, end-to-end speech synthesis systems have not been widely deployed in practical applications. In DurIAN, we replace the attention mechanism with an alignment model \cite{BLSTMWORLD,zen2016fast}, in which the alignment between the phoneme sequence and the target acoustic sequence is inferred from a phoneme duration prediction model. The duration of each phoneme is measured by the number of aligned acoustic frames. During training, the alignment between the acoustic frame sequence and the input phoneme sequence can be obtained through forced alignment widely used in speech recognition. The alignment is then used for hidden state expansion, which simply replicates hidden states according to phoneme duration. During synthesis, a separate duration model is exploited to predict the duration of each phoneme. This duration model is trained to minimize the mean squared error between the predicted phoneme duration and the duration obtained through forced alignment, given the whole sentence. After state expansion, the relative position of every frame inside each phone is encoded as a value between 0 and 1, and appended to the encoder state. The expanded encoder states are analogous to the attention context estimated in the end-to-end system, except that in DurIAN they are inferred from the predicted phone duration.

The duration model used in DurIAN is similar to the ones used in the conventional statistical synthesis models. It consists of three 512-unit bidirectional LSTM layers. Similar to that in the skip encoder, the states associated with the prosodic boundaries are also skipped before the final fully connected layer. 

\subsection{Decoder}
The decoder used in DurIAN is similar to the one used in Tacotron 1 \cite{wang2017tacotron}. The only difference is that the attention context concatenated with the \textit{pre-net} output is replaced with the encoder states derived from the alignment model in DurIAN. As in Tacotron, the decoder network can output single frame or multiple non-overlapped frames at each time step. When the target is multiple non-overlapped frames, a restricted attention is applied to the encoder states aligned with the target frames, and then concatenated with the output of the \textit{pre-net} at each time step. The attention mechanism used in DurIAN is different from that used in the end-to-end systems. In DurIAN, the attention context is computed from a small number of encoder frames that are aligned with the target frames. As long as the number of frames per decoder time step is not extraordinary large, it will not cause the similar artifacts observed in the end-to-end systems. The content-based tanh attention \cite{vinyals2015grammar} is used in our system and dropout with a probability of 0.5 is applied to the \textit{pre-net} in the decoder network during both training and inference. 

\subsection{Multimodal Synthesis}
The DurIAN is a generic framework that can be used for both speech and facial expression generation. The target for speech generation is mel-spectrogram and the target for facial expression generation is facial modeling parameters. The synchronization between the speech and the facial expression can be achieved through two different ways: either through multi-task learning, or through the duration model. When the multi-task learning based synchronization is used, the parallel speech and face data 
is required and the training target is the concatenation of the mel-spectrogram and the facial modeling parameters. However, when the duration-model based synchronization is used, the parallel speech and face data is no longer necessary since we can train two separate models independently, one for speech synthesis and one for facial expression generation, both share the same duration model trained with the speech alignment. In this study, we use duration based synchronization as it is more flexible and can pair different voice with different faces. 

\section{Fine-grained Style Control}\label{section:Style}
Fine-grained style control has recently been demonstrated in several unsupervised style disentangling approaches such as global style token (GST) \cite{wang2018style} and unsupervised variational autoencoder (VAE) \cite{hsu2018hierarchical,hsu2019disentangling}. As the objective of unsupervised style disentanglement is to disentangle the representations of speech into latent variables associated with speech generation, fine-grained control of spoken styles can be achieved by directly manipulating latent variable in continuous space. However, the challenge of unsupervised style disentanglement in practice is on the interpretation of the latent variables. As the latent variable is learned in unsupervised manner, it is difficult to uncover the 
physical interpretation of each latent variable. Moreover, the generation of speech with certain styles requires appropriate manipulation of several correlated latent variables simultaneously. For example, if you want to generate speech with a style of Sports Commentator, both the latent variables related to fundamental frequency, speaking rate, and other factors have to be controlled simultaneously. Uncovering the correlation between desired speaking styles and latent variables can be very difficult in the unsupervised framework. 

\begin{figure}[t!]
 \centering
  \includegraphics[clip, trim=0.2cm 12cm 0.2cm 12cm, width=1.0\textwidth]{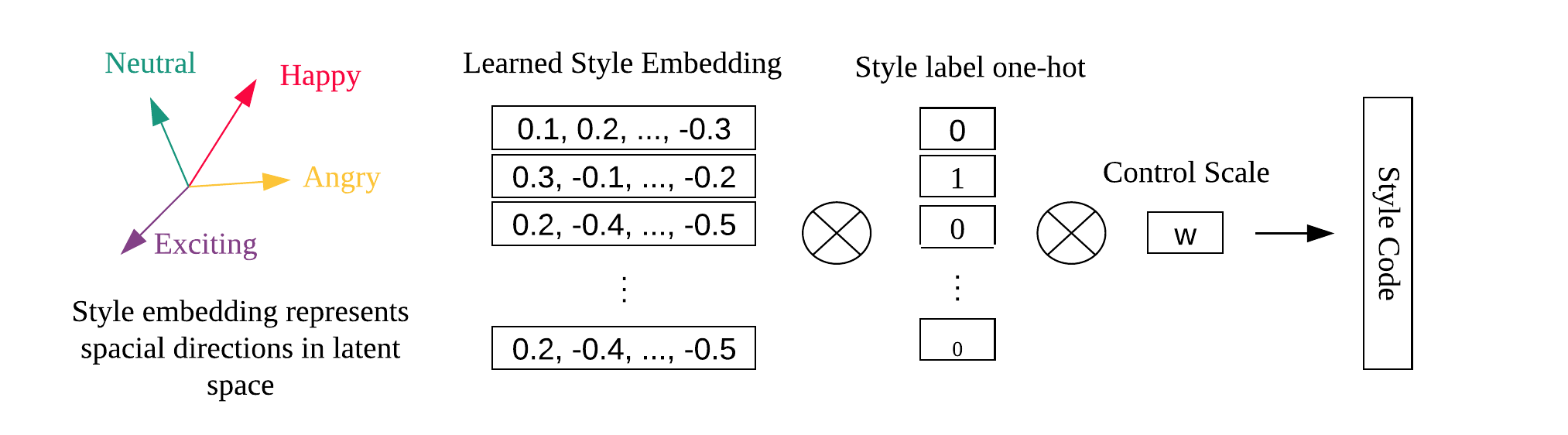}
 \caption{Generation of the style code from style embedding. The style embedding is learned jointly with other parameters of DurIAN through multi-style training.}
 \label{fig:StyleControl}
\end{figure}

On the other hand, fine-grained style control with supervised style labels has not been intensively investigated. In this section, we present a simple yet effective fine-grained style control algorithm in DurIAN. Here we assume only discrete style labels are available during training since fine-grained labels are difficult to obtain. Under this setup, while it is easy for TTS systems to generate speech in discrete styles indicated by the supervision labels in the training data, it is unclear how to extend the discrete labels into fine-grained style control during speech generation. In DurIAN, the fine-grained style control is achieved through the style code, a scaled version of style embedding. Figure.~\ref{fig:StyleControl} illustrates the process of generating the style code. The presumption behind scaling the learned style embedding is that the learned style embedding can be considered as a vector in the latent space where the direction can be interpreted as the attributes of the style and the magnitudes as the intensity of the corresponding style. Therefore, a fine-grained control of the style can be achieved by changing the magnitude of the learned style embedding without changing its direction. The style code is inserted in two different places in DurIAN as illustrated in Figure.~\ref{fig:DurIAN}. Since the duration of each phoneme is correlated with the speaking styles, the style code is exploited in the phoneme duration prediction model. Specifically, it is concatenated with the embedded representation of phoneme sequences. In addition, the style code is concatenated with the hidden states after the skip encoder to control the generation of acoustic features. The control scale for generating style code is set to a constant number of 1.0 during training, but can be any continuous value during inference to achieve fine-grained style control.

\section{Multi-band WaveRNN}
\label{sec:fastwavernn}

 \begin{figure}[t!]
 \centering
  \includegraphics[clip, trim=0.3cm 2.5cm 0.3cm 2.5cm, width=1.0\textwidth]{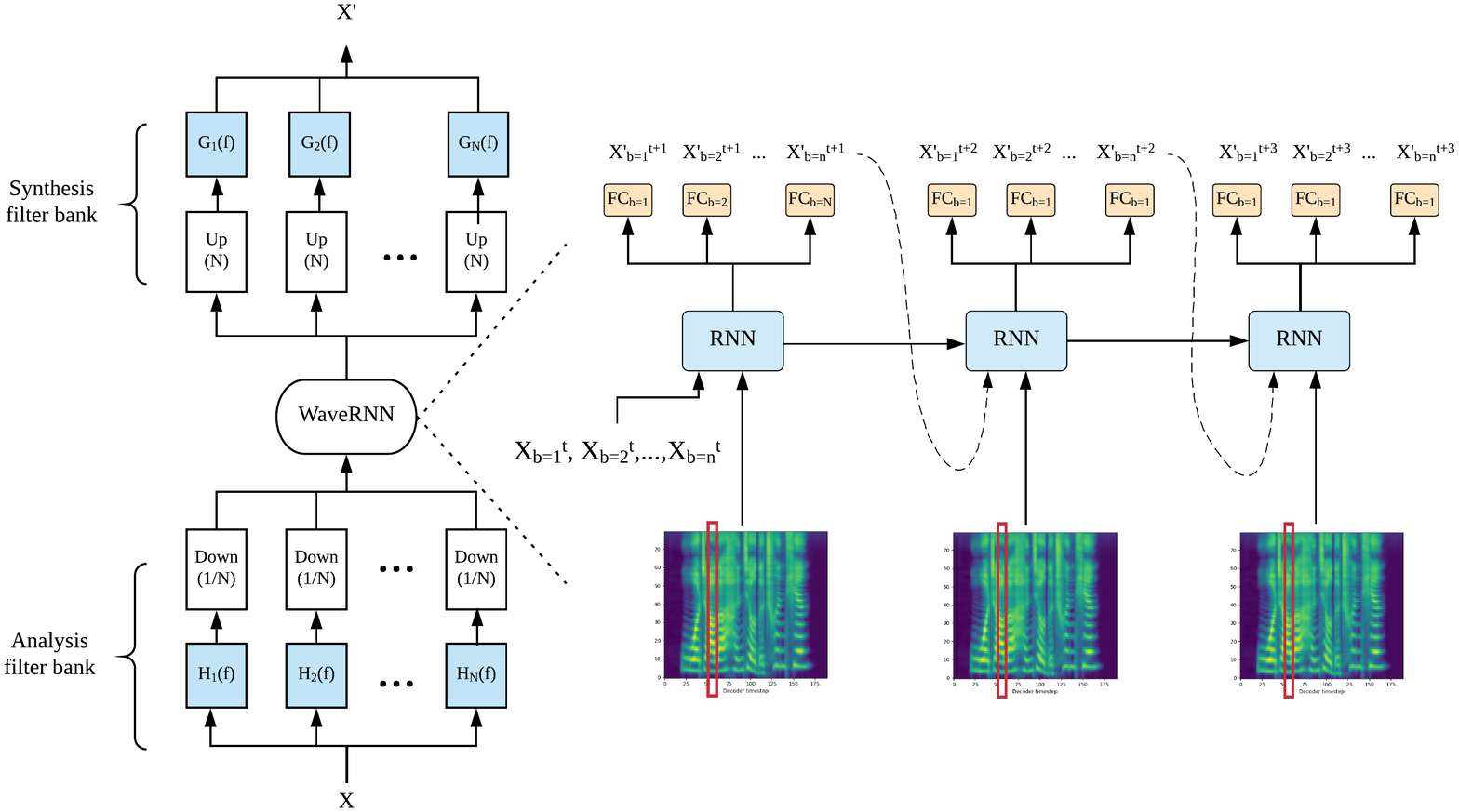}
 \caption{Model architecture of Multi-band WaveRNN.}
 \label{fig:FastWaveRNN}
 \end{figure}

Neural vocoders \cite{oord2016wavenet, oord2017parallel, kalchbrenner2018efficient, valin2019lpcnet} are capable of achieving highly natural speech that significantly surpasses that generated with the conventional vocoders. However, the major challenge of applying neural vocoders to production systems is the computational cost during inference. Most neural vocoders are designed to predict audio signals sample by sample. Therefore even one second of speech/audio requires tens of thousands of inference steps which significantly slows down the audio generation speed. In this work, we propose a multi-band synchronized parallel generation approach on top of the WaveRNN model to improve the inference speed of WaveRNN.

\subsection{WaveRNN}
The WaveRNN model we use follows the structure\footnote{The baseline WaveRNN model without weight matrix sparsification is used.} in \cite{kalchbrenner2018efficient}. A single-layer recurrent network and a dual-softmax layers is used to generate 16-bit audio. 
To accelerate WaveRNN inference speed, we performed 8-bit quantization on the hidden-layer weights of Gated recurrent units (GRU) as well as other four fully connected layers following it. The quantization significantly reduces the model size, which is very helpful to increase the cache hit rate. Moreover, the quantized parameters help to accelerate the calculations using the avx2 instruction of the Intel CPU. The combination of quantization and avx2 instruction could achieve 4x faster inference speed than floating-point calculations. Since direct quantization of the network causes the deterioration of synthesized sound quality, we use a quantitative loss learning mechanism during training to minimize the deterioration caused by quantization.

\subsection{Multi-band Parallel Strategy}
To further improve the audio generation speed, we propose the Multiband-WaveRNN alogorithm which can effectively reduce the computational cost. While several studies \cite{okamoto2018improving,okamoto2018investigation} have explored the usage of multi-band strategy for neural vocoder, these studies aim at modeling audio with higher sampling rate by training separate neural vocoder models for each subband. They require neural vocoders designed for different subband work in parallel in mulitple CPUs or GPUs. As a result, the total computational cost is not reduced.  In contrast, our proposed multi-band WaveRNN algorithm exploits the sparseness of the neural network model and uses a single shared WaveRNN model for all subband signal predictions. More specifically, the shared WaveRNN model takes all subband samples predicted from the previous step as input and predicts next samples in all subbands in one inference step as illustrated in Figure.~\ref{fig:FastWaveRNN}. We modify the original WaveRNN model to take inputs from multiple subbands and predict samples for all subbands simultanously through multiple output (and softmax) layers. With the proposed model structure, the audio in each subband can be downsampled by a factor of N (the number of frequency bands), and thus the total computational cost can be reduced. The predicted audio signals in each frequency band are upsampled first and then passed to synthesis filters. The signals from each frequency band after synthesis filter are summed to single audio signal. Downsamping is achieved by taking every Nth samples, and upsampling is done by padding zeros between original signal. The details of analysis and synthesis filters are described in Appendix~\ref{Appendix:filter design}.

\section{Experiments}
The performance of the proposed system is evaluated on three different tasks, including (1) speech synthesis, (2) fine-grained style control for expressive speech synthesis, and (3) multimodal speech synthesis.  

\subsection{Speech Synthesis}
\label{sec:exp_speech_synthesis}
We evaluated the naturalness and robustness of the proposed system using two different datasets. The first dataset is based on audio recordings from a professional male speaker. The training data contains a total of 18 hours of speech. Another dataset is from a professional female speaker and contains 7 hours of speech. Both subjects are native Mandarin speakers. All the training data has a sampling rate of 16KHz. 

Mean Opinion Score (MOS) of the naturalness of generated speech utterances are rated by human subjects participated in the listening tests. Two independent evaluations were performed using the models trained on male and female speakers, respectively. We use 40 unseen  sentences for evaluating the models trained from the male speaker, and 20 relatively longer out-of-domain sentences for evaluating the models trained from the female speaker. In all the experiments, 20 native Mandarin speakers participated in the listening test. We compared our model with the traditional BLSTM-based parametric system \cite{BLSTMWORLD} and the Tacotron-2 system. As shown in Table.~\ref{tab:MOS}, DurIAN and Tacotron~2 perform significantly better than the traditional parametric system. In both tests DurIAN and Tacotron-2 perform on-par with each other. No statistically significant difference can be observed. These results tell us that the superior naturalness in Tacotron-2 is likely a result of all other components in Tacotron other than the end-to-end attention mechanism. 

\begin{table}[h!]
\caption{5-scale mean opinion score evaluation.}
\centering
\begin{tabular}{@{}lll@{}}
\toprule
           & Male & Female \\ \midrule
Parametric & 3.54 & 3.47   \\
Tacotron 2 & 4.10 & 4.28   \\
DurIAN     & 4.11 & 4.26   \\ \bottomrule
\end{tabular}
\label{tab:MOS}
\end{table}

As the design goal of DurIAN is to achieve the naturalness comparable to Tacotron~2 while avoiding the artifacts observed in the Tacotron~2 system, We further compared two systems in robustness of generated speech. In this evaluation, we mainly focused on the word skipping and repeating errors commonly occur in the Tacotron~2 systems. Both DurIAN and Tacotron~2 systems were used to generate 1000 unseen utterances. The occurrence rate of word skipping and repeating errors are listed in Table.~\ref{tab:bad}. These results clearly indicate that DurIAN is much more robust than Tacotron-2 and generated no error in this category. 

\begin{table}[h!]
\caption{The occurrence rate of word skipping or repeating errors.}
\centering
\begin{tabular}{@{}lc@{}}
\toprule
           & skip/repeat \\ \midrule
Tacotron 2 \cite{shen2018natural} & 1\%          \\
Deep Voice 3 \cite{ping2017deep} & 4\%           \\ \midrule
Tacotron 2 & 2\%          \\
DurIAN     & 0\%           \\ \bottomrule
\end{tabular}
\label{tab:bad}
\end{table}

\subsection{Multi-band WaveRNN}
We evaluated the naturalness of generated speech and the speed of the Multi-band WaveRNN. 

\subsubsection{Speed}
The main complexity of WaveRNN comes from two GRUs and four fully-connected layers. We ignore the overhead of additive operations and focus only on the complexity of multiplication operations for each sample generated, which is

\begin{equation}
\mathrm{C} = 2*(\mathrm{2*\mathrm{N_{G}}*\mathrm{N_{G}}*3} +\mathrm{N_{G}}*\mathrm{N_{F}}+256*\mathrm{N_{G}*\mathrm{N_{B}}})*\mathrm{S_{R}}/\mathrm{N_{B}}, 
\end{equation}

where $\mathrm{N_{G}}$ is the size of the two GRUs, $\mathrm{N_{F}}$ is the width of
affine layer connected with final fully-connected layer,  $\mathrm{N_{B}}$ is the number of frequency band, and $\mathrm{S_{R}}$ is the sampling rate. Using  $\mathrm{N_{G}}=192$, $\mathrm{N_{F}}=192$ and $\mathrm{N_{B}}=1$ (fullband WaveRNN) for $\mathrm{S_{R}} = 16000$, we obtain a total complexity around 9.8 GFLOPS. When we set $\mathrm{N_{B}}=4$, the  total complexity is 3.6 GFLOPS. 

We also measured the Real Time Factor (RTF) for Multi-band WaveRNN systems listed in Table~\ref{tab:wavernnspeed}. All the RTF values were measured on a single Intel Xeon CPU E5-2680 v4 core. The results show that with quantization and avx2 speedup, the RTF can be reduced from 1.337 to 0.387 for the baseline WaveRNN model. With the 4-band model, the RTF can be further reduced to 0.171, which is 2x times faster than quantized WaveRNN model.

\begin{table}[h]
\caption{Real Time Factor (RTF) evaluation of proposed Multiband WaveRNN.}
\centering
\begin{tabular}{@{}lll@{}}
\toprule
RTF   & fullband & 4band \\ \midrule
float & 1.337    & 0.503 \\
int8  & 0.387    & 0.171 \\ \bottomrule
\end{tabular}
\label{tab:wavernnspeed}
\end{table}

\subsubsection{Quality}
The Mean Opinion Scores (MOSs) of proposed multi-band WaveRNN were obtained through subjective listening tests.
The female dataset used in Sec.~\ref{sec:exp_speech_synthesis} was used for training both the DurIAN and WaveRNN models. Three WaveRNN systems, the baseline WaveRNN model without quantization and the 4-band WaveRNN model with and without quantization, were compared. Experimental results in Table.~\ref{tab:wavernn_mos} indicate that the three systems evaluated are on-par with each other. No statistically significant difference was observed. If fact, most of the subjects participated in the listening tests cannot feel any difference between utterances generated from these three different WaveRNN systems. We can conclude that the proposed multi-band synthesis approach and the 8-bit quantization technique can effectively reduce the computational cost without deteriorating the quality of the generated speech.
\begin{table}[h!]
\caption{5-scale mean opinion score (MOS) evaluation of the proposed Multi-band WaveRNN.}
\centering
\begin{tabular}{@{}lc@{}}
\toprule
Systems                  & MOS  \\ \midrule
Fullband WaveRNN (float) & 4.53 \\
4-band WaveRNN (int8) & 4.58 \\
4-band WaveRNN (int8) & 4.56 \\ \bottomrule
\end{tabular}
\label{tab:wavernn_mos}
\end{table}

\subsection{Style Control}
In this experiment, we demonstrate the effectiveness of the proposed fine-grained style control approach for generating speech with different style scales. The corpus we used for experiments contains male speech collected for game commentary generation. It contains 4-hours of speech, of which 30-minutes of utterances are labeled with the \textit{exciting} style, 1-hour of utterances are labeled with the \textit{commentary} style, and the rest are marked as \textit{normal}. We generated game commentary samples with different levels of \textit{excitement} and a 5-minute video of real-time game commentary speech with DurIAN and our proposed fine-grained style control approach. The audio samples and the demonstration video can be found at https://tencent-ailab.github.io/durian.

\subsection{Multi-modal Synthesis}

In this experiment, we demonstrate the generation of synchronized speech and facial expression in 3D cartoon avatar generation task. For modeling facial expression, 32 dimensional fhw32b feature (face warehouse) were extracted from the 3D individual-specific blendshape collected from real person during training. Among the 32 dimensional fhw32b features, the first 25 dimensional features were extracted from the regression to the Principle Component Analysis(PCA) leading factors of face RGB features as described in \cite{mm1}. And the last 7 dimensional features record the 3D position of current facial expression. During the reconstruction process, the 3D avatar facial movement can be reconstructed \cite{mm2,mm3} using the fhw32b 3D facial feature predicted using DurIAN model. The demos can be found at https://tencent-ailab.github.io/durian.

\section{Conclusions}
In this paper, we presented a generic and flexible multimodal synthesis framework that is capable of generating highly natural and robust speech and facial expression. Our experimental results indicate that the proposed DurIAN system could synthesize speech with the naturalness and quality on par with the current state of the art end-to-end system Tacotron~2, at the same time effectively avoid the word skipping and repeating errors in generated speech. We also demonstrated a simple yet effective fine-grained style-control approach which controls not only the style but also the scale of the style of the generated speech. We have further demonstrated that using the proposed multi-band waveRNN we can speedup waveRNN inference time by at least two times over the already extensively optimized system without deteriorating the quality of generated speech. 

\section{Acknowledgments}
The authors would like to thank Drs. Linchao Bao, Haozhi Huang and other members in the Tencent AI Lab computer vision team for providing facial modeling features and multimodal experiment environment.

\bibliography{references}
\bibliographystyle{ieeetr}

\clearpage
\newpage
\section*{\Large Appendix}
\appendix
\section{Filter Design For Multiband WaveRNN}
\label{Appendix:filter design}
\begin{figure}[htbp]
\centering
\includegraphics[scale=0.8]{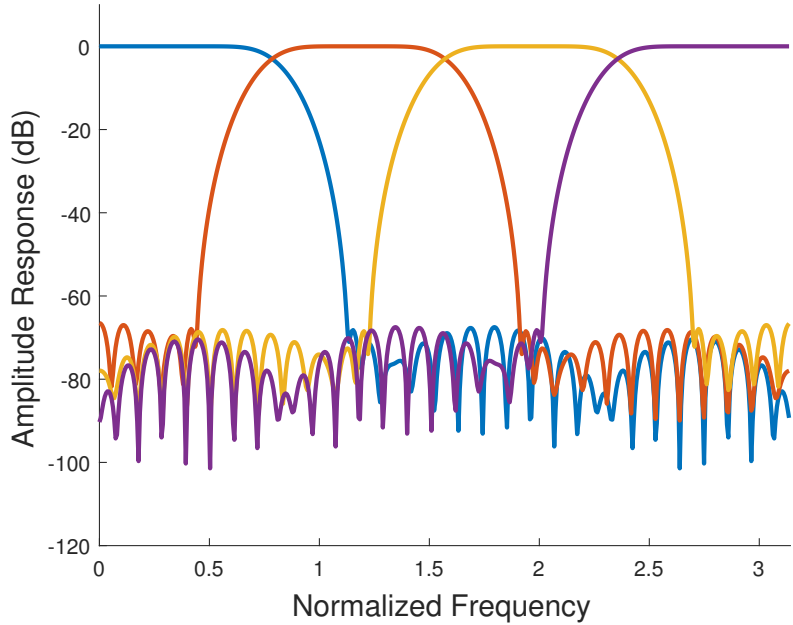}
\caption{Frequency response of 4-band  Pseudo-QMF filter banks. }
\label{fig:fbresp}
\end{figure}

A stable yet more efficient low cost filter bank, called Pseudo Quadratue Mirror Filter
Bank (Pseudo-QMF), is employed for our multi-band processing. Pseudo-QMF is a type of cosine-modulated filter bank (CMFB) where all the filters are cosine modulated version of a low-pass prototype filter. The prototype filter is designed to have a linear phase, leading to a phase-distortion-free analysis/synthesis system. Due to the aliasing cancellation constraint of the desired filer bank, the output aliasing is at the stopband attenuation level. As a result, by following the filter design in \cite{Nguyen94}, a high stopband attenuation property of analysis filter and synthesis filter can be achieved on an order of -100dB. Taking account of the computational efficiency of the multi-band processing, we choose the finite impulse response (FIR) analysis/synthesis filter order to of 63 for uniformly spaced 4-band implementations. It brings the stopband attenuation to the level of -70dB. The filter order is significantly smaller compared to the 1536 samples for analysis and synthesis FIR prototype filters realized in \cite{Okamoto18}. The frequency responses of 4-band uniform filter banks are plotted in Figure \ref{fig:fbresp}. The sampling frequency
of each subband waveform is $f_s / N$ Hz, where $N$ is the number of filter channels/subbands and $f_s$ is the desired sampling rate for the fullband signal. With the property of aliasing cancellation of Pseudo-QMF, the critical downsampling is applied after the fullband signal is decomposed into subbands by the analysis filterbank. In the future work, we will consider to use infinite impulse response (IIR) filter banks \cite{Crociere83} for even lower computational cost. 

\end{document}